\documentclass{article}
\usepackage{arxiv}
\usepackage[utf8]{inputenc} %
\usepackage[T1]{fontenc}    %
\usepackage[hidelinks]{hyperref}       %
\usepackage{url}            %
\usepackage{booktabs}       %
\usepackage{amsfonts}       %
\usepackage{nicefrac}       %
\usepackage{microtype}      %
\usepackage{lipsum}		%
\usepackage{graphicx}
\usepackage{threeparttable}
\usepackage{doi}
\usepackage{multirow}
\usepackage{multicol}
\usepackage{xcolor}
\usepackage{subcaption}
\usepackage[normalem]{ulem}
\usepackage{amsmath,amsfonts,bm}
\usepackage{amssymb,amsthm}
\usepackage{enumitem}
\usepackage{adjustbox}
\usepackage{blindtext}
\usepackage{cleveref}
\usepackage{float}
\usepackage{booktabs}
\usepackage{multirow}
\usepackage{array}
\usepackage{makecell} %
\usepackage[perpage]{footmisc}
\usepackage{wrapfig}

\usepackage[square,numbers,sort&compress]{natbib}
\setlist[itemize,1]{leftmargin=\dimexpr 18pt}
\setlist[enumerate,1]{leftmargin=\dimexpr 18pt}

\title{
\raisebox{-0.18\height}{\includegraphics[width=0.045\textwidth]{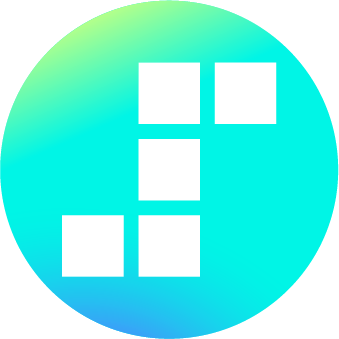}} %
Mind-Paced Speaking: A Dual-Brain Approach to Real-Time Reasoning in Spoken Language Models
}
\author{
Donghang Wu$^{1,2,*}$ \ Haoyang Zhang$^{1,2,}$\thanks{Equal contribution} \ Jun Chen$^{1}$ \ Xiangyu Tony Zhang$^{1,3}$ \ \textbf{Hexin Liu$^{2}$} \\ \textbf{Eng Siong Chng$^{2}$} \ \textbf{Fei Tian$^{1,}$\thanks{corresponding authors: \texttt{tianfei@stepfun.com}}}\quad \textbf{Xuerui Yang$^{1}$} \ \textbf{Xiangyu Zhang$^{1}$} \ \textbf{Daxin Jiang$^{1}$} \ \textbf{Gang Yu$^{1}$} \\
$^1$StepFun\quad $^2$ Nanyang Technological University \quad $^3$ University of New South Wales}

\renewcommand{\headeright}{Preprint. Work in progress.}
\renewcommand{\undertitle}{}
\renewcommand{\shorttitle}{}

\begin{document}
\large

\maketitle
\begin{abstract}
Real-time Spoken Language Models (SLMs) struggle to leverage Chain-of-Thought (CoT) reasoning due to the prohibitive latency of generating the entire thought process sequentially. Enabling SLMs to think while speaking, similar to humans, is attracting increasing attention. We present, for the first time, Mind-Paced Speaking (MPS), a brain-inspired framework that enables high-fidelity, real-time reasoning. Similar to how humans utilize distinct brain regions for thinking and responding, we propose a novel dual-brain approach, employing a ``Formulation Brain'' for high-level reasoning to pace and guide a separate ``Articulation Brain'' for fluent speech generation. This division of labor eliminates mode-switching, preserving the integrity of the reasoning process. Experiments show that MPS significantly outperforms existing think-while-speaking methods and achieves reasoning performance comparable to models that pre-compute the full CoT before speaking, while drastically reducing latency. Under a zero-latency configuration, the proposed method achieves an accuracy of 92.8\% on the mathematical reasoning task Spoken-MQA and attains a score of 82.5 on the speech conversation task URO-Bench. MPS is the methodology underlying our released \textbf{Step-Audio R1.1} system, effectively bridging the gap between high-quality reasoning and real-time interaction\footnote[1]{Code: \href{https://github.com/stepfun-ai/Step-Audio-R1\#overview-of-r11}{\nolinkurl{github.com/stepfun-ai/Step-Audio-R1}}~\textbar~Model: \href{https://huggingface.co/stepfun-ai/Step-Audio-R1.1}{\nolinkurl{huggingface.co/stepfun-ai/Step-Audio-R1.1}}}.
\end{abstract}

\enlargethispage{2.5\baselineskip}
\vspace{4pt}
\begingroup
\centering
\includegraphics[width=0.92\textwidth]{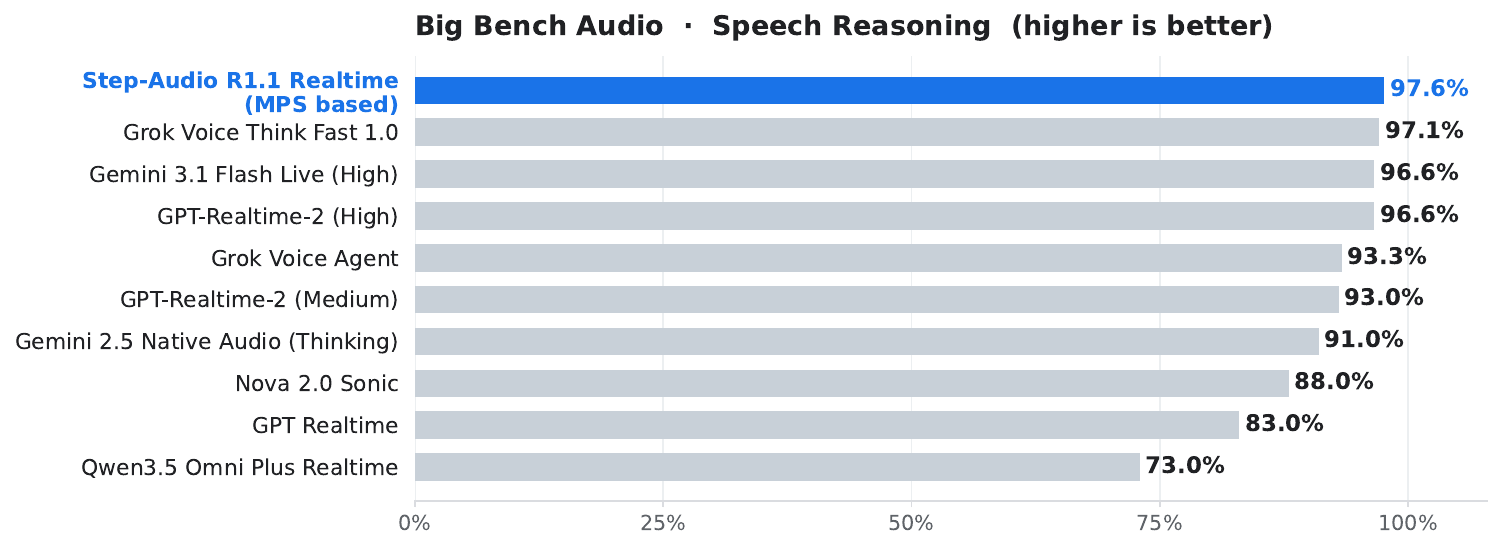}\par
\vspace{2pt}
\refstepcounter{figure}\label{fig:bigbench_audio}%
{\small\noindent\textbf{Figure~\thefigure:} Speech reasoning accuracy on the Artificial Analysis Big Bench Audio benchmark~\citep{aaspeechleaderboard}. Step-Audio R1.1 Realtime, built on the proposed Mind-Paced Speaking (MPS) framework, performs comparably to or above frontier speech-to-speech systems~\citep{geminiflashlive}.\par}
\endgroup

\clearpage

\section{Introduction}
Speech has emerged as a more natural and fundamental modality for human-computer interaction, leading to growing emphasis on spoken language models (SLMs) \cite{slmsurvey,stepaudio2,salmduplex,salmonn,moshi}. These models facilitate seamless communication by processing and generating audio-based inputs and outputs. A key component enhancing their capability is the integration of thinking, particularly through Chain-of-Thought (CoT) processes and its extensions \cite{cot,selfconsistency,tree_of_thought,pal}, as implemented in frameworks like Think-Before-Speak (TBS) \cite{cot,tbs1,tbs2}. This approach enables models to decompose complex tasks into step-by-step reasoning sequences, thereby improving interpretability and performance in dialogue systems.

However, generating complete CoT sequences often introduces significant latency, which hinders real-time applications. Recent efforts to reduce reasoning latency have garnered significant attention \cite{stitch,miniomnireasoner}. These methods explore "think-while-speaking" paradigms, where models interleave thinking and response tokens. The Large Language Model (LLM) continuously switches between think and response modes. It first generates several think tokens, then produces several response tokens based on them. These response tokens are sent to the Text To Speech (TTS) system for speech synthesis. While the speech is synthesizing, the LLM continues to generate more think tokens. However, this interleaving disrupts semantic coherence by forcing the model to frequently switch between thinking and response generation, potentially degrading the performance.

In fact, the human brain provides a biological analogy for efficient parallel processing. Cognitive neuroscience reveals that thinking and speaking involve distinct brain areas \cite{Cognitivebasis, cortical}. Speech does not follow a rigid "think-then-speak" sequence or interleaved sequence. Crucially, it exhibits an incremental nature where later parts of a thought are still being processed while the initial parts of the utterance are already being spoken \cite{thkwhilespk}. 
Inspired by this, we introduce \textbf{M}ind-\textbf{P}aced \textbf{S}peaking (MPS), a novel architecture for enabling SLMs to "think" and "speak" in a concurrent and integrated manner. The core of MPS is a dual-brain framework that operates analogously to the human cognitive-speech system. One LLM acts as a central "Formulation Brain", continuously generating an internal stream of thought. The other functions as an Articulation Brain, which receives this thought stream in segments and generates the corresponding spoken output. The Formulation Brain does not need to complete a full reasoning chain before the Articulation Brain begins. Instead, the ongoing thinking process actively sets the pace and provides the contextual guidance for the Articulation Brain, allowing it to vocalize fluently even as the underlying thoughts are still being formed and refined by the Formulation Brain. This mind-paced mechanism ensures that the spoken output is not only grounded in a thinking process but also maintains semantic coherence, closely mimicking the natural human process of thinking while speaking. Furthermore, we propose a think-incomplete Supervised Fine-Tuning (SFT) method to enable the Articulation Brain to respond based on incomplete thinking content. The experimental results on benchmarks such as mathematical reasoning, dialogue, and question-answering, prove that compared to methods that answer directly without thinking, or existing methods that think while speaking, the proposed MPS method effectively utilizes the thinking process and continuous semantic context, obtaining more accurate and higher-quality responses. Compared to TBS method, the proposed MPS significantly reduces response latency while maintaining performance.

Our main contribution can be summarized as follows:

(1) We propose an MPS architecture that enables SLMs to achieve human-like think-while-speaking capabilities. This method significantly reduces the latency of the CoT process while maintaining the semantic coherence of the LLM. Consequently, the LLM leverages the CoT content to deliver superior performance.

(2) We develop a think-incomplete SFT to train LLMs to generate responses based on partial thinking processes, thereby enabling them to perform think-while-speaking.

(3) We evaluate two distinct MPS architectures: Speak-First and Think-First, against baseline methods. Experimental results demonstrate that the proposed think-while-speaking MPS method significantly outperforms both the direct response approach without a thinking process and existing interleaved think-while-speaking methods. Compared to the TBS architecture, our method substantially reduces response latency while maintaining the quality of the LLM's responses.

(4) Our proposed MPS architecture mimics the neuroscientific mechanisms of human thinking and speaking, transcending the structural limitations of existing interleaved think-while-speaking methods. It provides the research community with a reference paradigm for subsequent studies on anthropomorphic, real-time dialogue systems. The proposed framework also serves as the methodology behind our released Step-Audio R1.1 system; see Figure~\ref{fig:bigbench_audio} for its standing on the Artificial Analysis Speech Reasoning leaderboard.

\section{Related Work}
\subsection{Spoken Language Models}
Spoken Language Models (SLMs) accept user speech input and generate speech output, enabling real-time speech dialogue with users. Since the LLM backbone is typically trained in the text domain, directly generating speech tokens presents a challenge \cite{stitch}. Most current SLMs first generate text tokens and then generate speech. This is achieved through two primary methods: one approach uses the LLM to output text, which is then synthesized into speech by an additional TTS model \cite{qwen2.5omni,llamaomni2}. Another method employs the LLM to generate interleaved text and audio tokens, where each output chunk contains a fixed number of text tokens and speech tokens, and a speech decoder directly synthesizes the speech signal \cite{glm,stepaudio2}. For example, Step-Audio 2 produces output in the \textit{ta4} format, which means it outputs one text token followed by four audio tokens. This chunk of output is then passed through a speech detokenizer to obtain the speech signal \cite{stepaudio2}.

\subsection{Reason for SLMs}
Although explicit CoT has been proven helpful in text LLMs, most SLMs still lack CoT capability. One reason is that audio and text have different structures; another reason is that directly synthesizing CoT into speech increases the confusion of responses, while generating silent CoT introduces significant latency, which becomes unreasonable in daily conversations. Some studies introduce the reasoning ability into Audio LLMs \cite{stitch}. For example, Xie et al. have proposed an audio CoT reasoning dataset to fine-tune models \cite{audioreasoner}. Some studies use reinforcement learning, such as GRPO \cite{grpo}, to fine-tune models and enhance their reasoning ability \cite{rlreason1,rlreason2}. However, these studies remain limited to audio-in-text-out Audio LLMs, not SLMs that can engage in dialogue with humans. In \cite{stepaudio2}, Step-Audio 2, which takes speech as input and output, using CoT and reinforcement learning to improve the response qualities, is proposed. Step-Audio 2 offers a solution for introducing explicit reasoning into SLMs. Some methods achieve simultaneous thinking and speaking by segmenting CoT content and response content, using the LLM to generate interleaved think tokens and response tokens \cite{stitch,miniomnireasoner}. However, this approach differs from the LLM's original response generation format. The LLM needs to continuously switch between think mode and response mode, which disrupts semantic coherence and affects its performance.

\section{Method}
This section first outlines the conventional TBS-based SLM. We then present the proposed MPS method. We also introduce the think-incomplete SFT, which is designed to teach LLMs the think-while-speaking capability.

\begin{figure}[t]
\begin{center}
\begin{minipage}[b]{0.89\linewidth}
  \centering
  \centerline{\includegraphics[width=14cm]{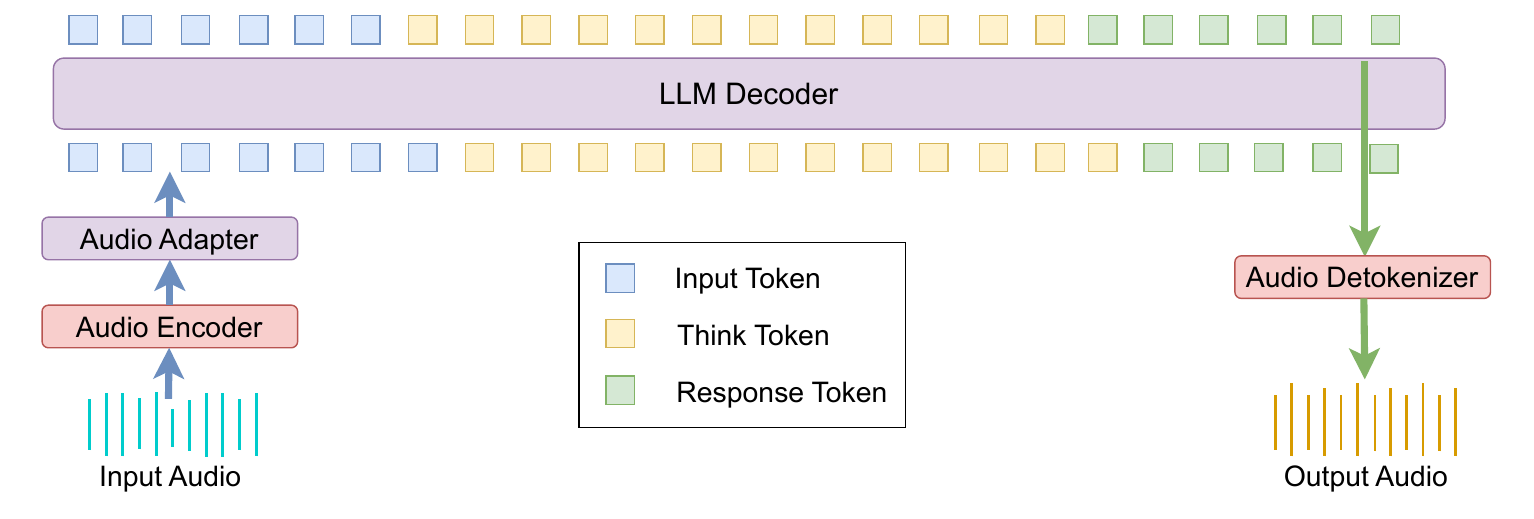}}
\end{minipage}
\vspace{-0.2cm}
\caption{Architecture of the TBS architecture. For the sake of conciseness, we remove the input text, which is optional in SLMs. The TBS SLM first generates the full CoT and then produces response tokens.}
\label{fig:tbs}
\end{center}
\end{figure}

\subsection{Think Before Speaking}
The architecture of TBS-based SLM is shown in Figure \ref{fig:tbs}. To enhance the reasoning ability of SLM, the TBS paradigm, after receiving user speech $\bm{X}^{\mathrm{spc}}$ and optional text instructions $\bm{X}^{\mathrm{txt}}$, first generates step-by-step CoT tokens $\bm{Y}^{\mathrm{cot}} \in \mathbb{R}^{T_c}$, and then generates the response tokens $\bm{Y}^{\mathrm{res}} \in \mathbb{R}^{T_r}$, where $T_c$ and $T_r$ denote the number of CoT tokens and response tokens, respectively. This can be divided into two processes: the thinking process and the speaking process. The thinking process can be written as:
\begin{equation}
\label{equa:think}
\begin{split}
    P_{\theta_l}(\bm{Y}^{\mathrm{cot}} | \langle\bm{X}^{\mathrm{spc}}, \bm{X}^{\mathrm{txt}}\rangle) = \prod_{t=1}^{T_c}P_{\theta_l}(Y^{\mathrm{cot}}_t|\langle \bm{Y}^{\mathrm{cot}}_{1:t-1}, \bm{X}^{\mathrm{spc}}, \bm{X}^{\mathrm{txt}}\rangle),
\end{split}
\end{equation}
where $\theta_{l}$ denotes the parameters of the SLM. After that, the LLM generates response tokens for speaking, which can be formulated as:
\begin{equation}
\label{equa:resp}
\begin{split}
    P_{\theta_l}(\bm{Y}^{\mathrm{res}} | \langle\bm{Y}^{\mathrm{cot}}, \bm{X}^{\mathrm{spc}},\bm{X}^{\mathrm{txt}}\rangle) = \prod_{t=1}^{T_r}P_{\theta_l}(Y^{\mathrm{res}}_t|\langle \bm{Y}^{\mathrm{res}}_{1:t-1}, \bm{Y}^{\mathrm{cot}}_{1:T_c}, \bm{X}^{\mathrm{spc}}, \bm{X}^{\mathrm{txt}}\rangle).
\end{split}
\end{equation}
Through this method, the task is decomposed into a step-by-step process. Additionally, by introducing CoT tokens, it enables more Transformer forward operations and thus gives LLM a deeper inference depth \cite{pause,dots}. 

\subsection{Architecture}
In the human brain, speech production is not a monolithic process but the result of two highly specialized and collaborative systems. The first, a network centered around the prefrontal-temporal cortex, is responsible for high-level cognitive functions such as conceptualization, logical reasoning, and content planning. Subsequently, a second system, primarily involving the motor cortex and subcortical pathways, translates these abstract thoughts into natural language for articulation, enabling fluent speech. These two systems operate in parallel, with the cognitive system continuously supplying thinking content to the articulatory system, creating a natural flow where the mind paces speech \cite{Cognitivebasis,cortical}.

\begin{figure}[t]
\begin{center}
\begin{minipage}[b]{0.8\linewidth}
  \centering
  \centerline{\includegraphics[width=17cm]{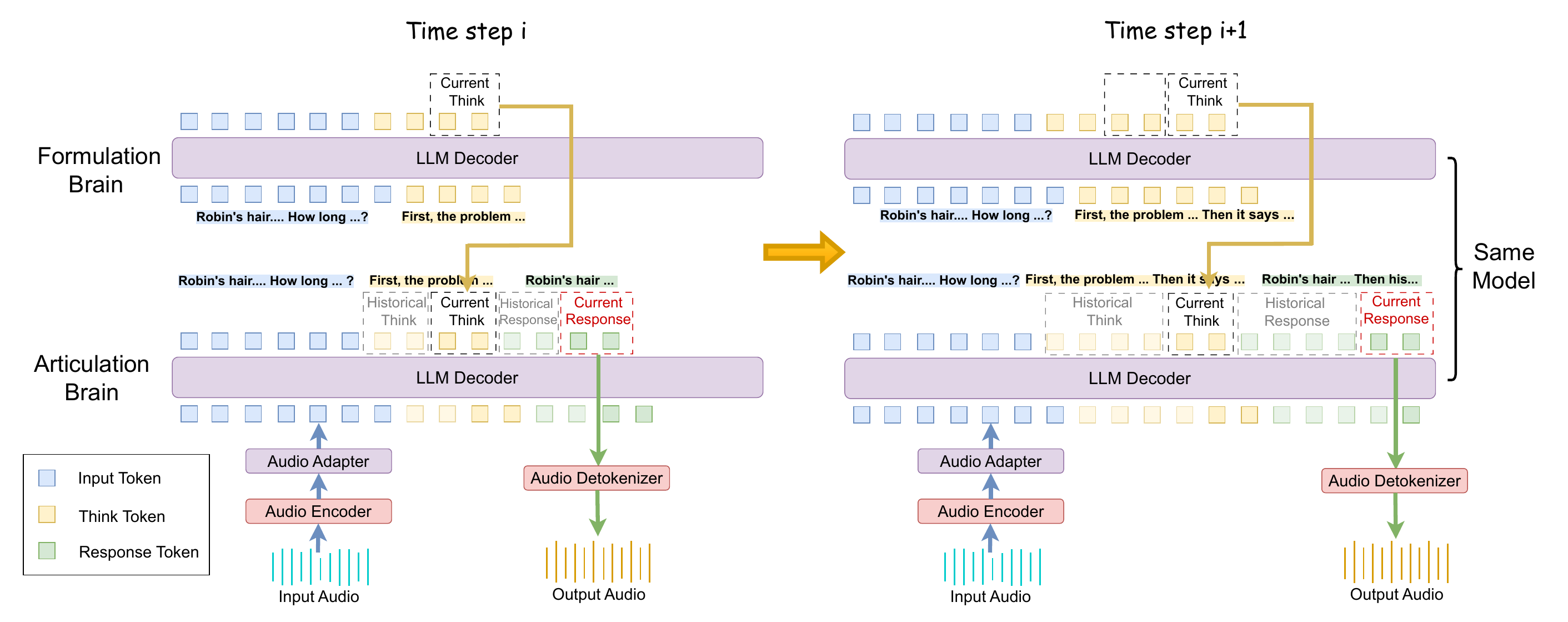}}
\end{minipage}
\caption{Architecture of the proposed MPS. For the sake of conciseness, we remove the input text, which is optional in SLMs. We demonstrate the process from step i to step i+1 when generating think segments and response segments. The Formulation Brain LLM continuously generates the think segments. The newly generated think segment and the response segment from the previous step are both added as the prefix to the Articulation Brain LLM, pacing the Articulation Brain LLM to produce response segment correspondingly.}
\label{fig:architecture}
\end{center}
\end{figure}

Inspired by this, we abstract this mechanism of separated "formulation" and "articulation" into our model architecture. Instead of relying on a single LLM to handle both thinking and speaking, we propose a dual-brain system composed of two distinct LLMs.

Our proposed framework, illustrated in Figure \ref{fig:architecture}, leverages a dual-LLM architecture consisting of a Formulation Brain LLM and an Articulation Brain LLM. The Formulation Brain LLM is dedicated to user intent understanding and performs deliberate CoT reasoning, with its internal process materialized as "think tokens". Subsequently, the Articulation Brain LLM converts this structured reasoning and the dialogue context into natural language, producing the final "response tokens" for spoken output.


\textbf{Formulation Brain}: The Formulation Brain's operating mode is identical to that of TBS Audio LLMs but with only the thinking process. After receiving user input $\bm{X}^{\mathrm{spc}}$ and $\bm{X}^{\mathrm{txt}}$, it aims to generate the step-by-step CoT tokens $\bm{Y}^{\mathrm{cot}} \in \mathbb{R}^{T_c}$. We use the tokens \texttt{<think>} and \texttt{</think>} to mark the beginning and end of the CoT. This process can be formulated as (\ref{equa:think}). In the MPS architecture, we do not wait for the Formulation Brain to complete the entire CoT before the Articulation Brain starts speaking. We divide the CoT tokens $\bm{Y}^{\mathrm{cot}}$ into N segments, denoted as $[\bm{S}^{\mathrm{cot}}_1, \bm{S}^{\mathrm{cot}}_2, ..., \bm{S}^{\mathrm{cot}}_N]$. Each time the Formulation Brain produces a think segment $\bm{S}^{\mathrm{cot}}_n$, we feed the segment to the Articulation Brain, which then generates a response segment based on the current think segment and historical thinking and response contents. After the Formulation Brain LLM finishes CoT segments, it stops generating response tokens as we do not require the Formulation Brain to speak.

\textbf{Articulation Brain}: The Articulation Brain accepts the same user input as the Formulation Brain. After obtaining the current think segment $\bm{S}^{\mathrm{cot}}_n$ from the Formulation Brain, we concatenate it with the historical think segments $[\bm{S}^{\mathrm{cot}}_1, \bm{S}^{\mathrm{cot}}_2, ..., \bm{S}^{\mathrm{cot}}_{n-1}]$, placing \texttt{<think>} and \texttt{</think>} at the beginning and the end, and then append the historical response segments, which are defined as $[\bm{S}^{\mathrm{res}}_1, \bm{S}^{\mathrm{res}}_2, ..., \bm{S}^{\mathrm{res}}_{n-1}]$. This allows the Articulation Brain to continue generating the subsequent response content. After that, we use a streaming TTS model to synthesize speech in real-time. The Articulation Brain's output is incremental. For every think segment that the Formulation Brain produces, the Articulation Brain generates a segment of the response $\bm{S}^{\mathrm{res}}_n$. This process can be written as: 
\begin{equation}
\begin{split}
    P_{\theta_l}(\bm{S}^{\mathrm{res}} | \langle \bm{S}^{cot}, \bm{X}^{\mathrm{spc}}, \bm{X}^{\mathrm{txt}}\rangle) = \prod_{n=1}^{N}P_{\theta_l}(\bm{S}^{\mathrm{res}}_n|\langle \bm{S}^{res}_{1:n-1}, \bm{S}^{cot}_{1:n}, \bm{X}^{\mathrm{spc}}, \bm{X}^{\mathrm{txt}}\rangle),
\end{split}
\end{equation}
When the Formulation Brain just begins its thinking, the Articulation Brain can only generate a response segment based on a small amount of CoT. The response segment it generates at this stage may be of lower quality. As the Formulation Brain's thinking content increases, the Articulation Brain receives more CoT content, and it subsequently generates responses of increasingly higher quality.


Compared with existing think-while-speaking methods that use a single LLM to predict interleaved think and response tokens, thereby forcibly interrupting and splitting the originally continuous think and response content \cite{stitch,miniomnireasoner}, our method adopts a dual-brain design consisting of the Formulation Brain and the Articulation Brain.
From the perspectives of the Formulation Brain and the Articulation Brain, both are classic TBS LLMs that, after receiving user input, first generate step-by-step CoT content and then generate response content conditioned on the CoT, thereby greatly ensuring the semantic coherence of the LLM output.
By allowing the Formulation Brain to pace the Articulation Brain, our method achieves a human-like think-while-speaking process.
\subsection{Think-incomplete SFT}
Since the proposed MPS method does not change the input-output patterns of the classic LLM for the individual Formulation Brain and Articulation Brain, the proposed MPS, unlike existing think-while-listening methods \cite{stitch,miniomnireasoner}, does not require repretraining the LLM. 
To ensure that the Articulation Brain LLM possesses the ability to accept incomplete think content and produce reasonable output, we introduce think-incomplete SFT. In the construction of training data, we randomly retain the content of the first $L$ steps of the step-by-step CoT, delete the subsequent CoT content, then place this incomplete CoT with \texttt{<think>} and \texttt{</think>} tokens at the beginning and end, concatenate it with the groundtruth response, and use it as the next-token-prediction training objective for the LLM \cite{ntp}.

During the inference stage, we use segments with a fixed number of tokens. We set $T_c$ and $T_r$ to $80$ and $100$ respectively. We use the output format of Step-Audio 2, specifically the \textit{ta4} format, which generates one text token followed by four speech tokens, thus every $100$ response tokens contain $20$ text tokens and $80$ speech tokens. We also attempt to use the same segment division strategy as in the think-incomplete SFT phase, but we find that it does not bring improvement; on the contrary, it introduces uncontrollable latency due to the variable length of each CoT step. We also try using a fixed token count strategy for dividing the CoT during the think-incomplete SFT phase, but it does not yield performance improvements either.
\section{Experiments}
\subsection{Experimental Settings}
\label{sec:latency}
The LLM backbone used in this paper is Step-Audio 2, and its parameter settings refer to \cite{stepaudio2}. The LLMs in Formulation Brain and Articulation brain share the same parameters. To verify the effectiveness of the proposed method on tasks requiring reasoning, we use Spoken-MQA, a mathematical reasoning dataset \cite{spokenmqa}. We use accuracy as the evaluation metric. Furthermore, to validate the method's effectiveness on general dialogue tasks, we introduce URO-Bench, which contains several subtasks such as daily dialogue, emotion recognition, paralinguistic information, and question-answering \cite{urobench}. For question-answering tasks, we use accuracy as the metric. For other tasks, we use GPT-score, generated by \texttt{GPT-4o-mini} and ranging from 0 to 100, to evaluate response quality.

To accommodate the latency requirements of different application scenarios, we implement two distinct MPS paradigms:
\begin{itemize}
\item Think-First, denoted as \textbf{MPS-thkfirst}: The Formulation Brain LLM first generates $T_c$ think tokens, after which the Articulation Brain LLM generates $T_r$ response tokens and synthesizes speech. Under this setting, the latency is $T_c$ plus the buffer size of streaming TTS, which is significantly lower than the latency required for the TBS structure to generate a complete Chain-of-Thought.

\item Speak-First, denoted as \textbf{MPS-spkfirst}: The Articulation Brain LLM first generates $T_r$ response tokens, while simultaneously, the Formulation Brain LLM begins generating think tokens. The Formulation Brain LLM completes generating $T_c$ think tokens before the speech synthesized from the $T_r$ response tokens finishes playing. In this configuration, 
the latency is solely the buffer size of the streaming TTS, meaning the model can be considered to respond directly with near-zero latency.
\end{itemize}

Additionally, we compare the proposed method with two approaches that use the same LLM backbone as in this paper: Think-Before-Speaking (MPS-tbs) and direct response without thinking (MPS-wo/\textit{thk}), to validate the effectiveness of our proposed think-while-speaking methodology. 

\subsection{Data Construction}
We begin with real-world user queries as our seed set. To ensure topical diversity and sufficient scale, we employ \texttt{GPT-4o} \cite{gpt4o} for transcription and augmentation of these queries. These augmented queries are then used as user prompts to distill dialogue data with native CoT from the \texttt{DeepSeek-R1} model \cite{tbs2}.

However, the raw data generated by \texttt{DeepSeek-R1}, a text-centric model, presents two critical challenges for spoken dialogue applications: (1) Text-specific stylizations, such as Markdown formatting and emojis, which are incompatible with speech synthesis. (2) The CoT data reflects complete, turn-based reasoning chains, a format unsuitable for training the model to respond from partial thoughts. When the CoT generated by the LLM exhibits some incomplete, its performance is affected.

To address these challenges, we implement a fine-grained data processing pipeline:
\begin{itemize}
 \item \textbf{Compatibility Processing:} We discard samples containing Markdown formatting or multi-item lists that cannot be naturally rendered in speech. For samples containing emojis, we employ \texttt{Qwen-72B-Instruct} \cite{qwen2} to remove these elements while preserving the plain text content.

 \item \textbf{CoT Pruning:} To train the model to respond stably with only partial CoT, we augment the data by randomly deleting some reasoning paragraphs. This operation is performed in a way that generally preserves the overall logic of the CoT. Crucially, to maintain the stylistic distribution of the original DeepSeek CoT, we neither delete individual sentences within a paragraph nor use an LLM to rewrite the content of the remaining parts. This ensures that the preserved paragraphs are stylistically and distributionally consistent with the source model.
\end{itemize}

\subsection{Results}
\begin{table}[t]
\caption{Test-set accuracy (\%) of different methods on the Spoken-MQA benchmark. The evaluated approaches include: the direct response baseline without a thinking process (MPS-wo/\textit{thk}), Think-Before-Speaking (MPS-tbs), Think-First (MPS-thkfirst), and Speak-First (MPS-spkfirst). Results of baseline systems are taken from \cite{miniomnireasoner} except that results of Step-Audio 2 are reproduced by ourselves.}
\label{tab:spokenmqa}
\renewcommand\arraystretch{1.1}
\begin{center}
\begin{tabular}{lccccccc}
\toprule
\multirow{2}{*}{Method} & 
  \multicolumn{3}{c}{Arithmetic} & 
  \multicolumn{3}{c}{Reasoning} & 
  \multirow{2}{*}{Avg} \\
  \cmidrule(lr){2-4} \cmidrule(lr){5-7}
  & Short & Long & Avg & Single & Multi & Avg
\\ \midrule
Whisper-Qwen2.5-7B-Instruct & - & - & 70.0 & - & - & 72.5 & 72.2\\
Whisper-Qwen2.5-Math-7B-Instruct & - & - & 77.3 & - & - & 86.7 & 85.6\\
\midrule
LLaMA-Omni & 40.0 & 11.0 & 23.5 & 29.5 & 10.5 & 16.2 & 16.8 \\
Mini-Omni & 5.0 & 2.3 & 3.5 & 0.8 & 1.9 & 1.6 & 1.7 \\
Freeze-omni & 43.0 & 14.5 & 26.8 & 69.0 & 19.8 & 34.4 & 33.3 \\
GLM-4-Voice & 40.0 & 22.5 & 30.1 & 54.4 & 28.5 & 36.2 & 35.3\\
Qwen2-Audio-7B-Instruct & 43.0 & 31.2 & 36.3 & 55.4 & 22.5 & 32.3 & 32.7 \\
Qwen2.5-Omni-7B & 83.0 & 45.1 & 61.5 & 85.2 & 71.5 & 75.6 & 73.6 \\
Qwen2.5-Omni-3B & 84.0 & 43.3 & 60.1 & 81.5 & 57.1 & 64.4 & 63.6\\
Mini-Omni-Reasoner & \textbf{92.9} & 66.1 & 77.3 & 85.9 & 60.5 & 68.1 & 68.6\\
Step-Audio 2 & 89.0 & 52.6 & 65.9 & 95.6 & 90.4 & 91.9 & 88.8\\
\midrule
MPS-wo/\textit{thk} & 71.0& 34.1& 47.6 & 88.0& 67.8& 73.8 & 70.6\\
MPS-tbs & 90.0 & \textbf{88.4} & \textbf{89.0} & 94.4 & 93.2& 93.6& 93.0\\
MPS-thkfirst & 89.0 & 84.9 & 86.4 & 95.6 & \textbf{94.6}& \textbf{94.9} & \textbf{93.9}\\
MPS-spkfirst & 87.0 & 71.7& 77.3 & \textbf{96.0}& 94.5& \textbf{94.9}& 92.8 \\
\bottomrule
\end{tabular}
\end{center}
\vspace{-10pt}
\end{table}
\subsubsection{Evaluation on Reasoning Tasks}

Table \ref{tab:spokenmqa} shows the computational accuracy on Spoken-MQA. It can be seen that the proposed MPS-thkfirst method exceeds the MPS-wo/thk method and all baseline methods, including the think-while-speaking Mini-Omni-Reasoner, in all evaluation tasks. The results proves that the proposed method effectively utilizes the thinking process, achieving more intelligent response. Compared to Mini-Omni-Reasoner, the MPS method maintains semantic coherence and achieves better performance.
\begin{table}[t]
\small
\caption{The average accuracy of different models with CoT capability on Spoken-MQA, and the extra tokens generated by the model before generating the first response token. The evaluated approaches include: Interleaved Think-While-Speaking (Mini-Omni-Reasoner), Think-Before-Speaking (MPS-tbs), Think-First (MPS-thkfirst), and Speak-First (MPS-spkfirst).}
\label{tab:latency}
\renewcommand\arraystretch{1.1}
\centering
\begin{tabular}{lcc}
\toprule
\multicolumn{1}{c}{Method}  &\multicolumn{1}{c}{Accuracy} &\multicolumn{1}{c}{Extra Tokens}
\\ \midrule
Mini-Omni-Reasoner & 68.6\% & 8 \\
MPS-tbs & 93.0\% & 762\\
MPS-thkfirst   & \textbf{93.9\%} & 80\\
MPS-spkfirst   & 92.8\% & \textbf{0}\\
\bottomrule
\end{tabular}
\end{table}
Besides, compared to MPS-tbs, the MPS-thkfirst method demonstrates comparable performance, being slightly weaker in arithmetic computation tasks but superior in reasoning tasks. One possible explanation is that reasoning tasks require more textual analysis. The MPS-thkfirst method, by using each think segment to pace the generation of a corresponding response segment, implicitly achieves semantic alignment, enabling the model to better utilize contextual information for response generation.

The MPS-spkfirst method experiences some performance degradation because it initially outputs a response segment without utilizing any think segments. This impact is particularly pronounced in tasks involving direct arithmetic computation. For reasoning tasks, experimental observations indicate that the initial phase of the LLM's CoT content primarily involves analyzing the semantic information of the question, often rewriting the question's content. Consequently, this initial portion of the CoT content has a limited effect on the final response. As a result, MPS-spkfirst is minimally affected in reasoning tasks, and its performance remains nearly identical to that of MPS-thkfirst. Experimental results on Spoken-MQA demonstrate that the proposed method significantly leverages CoT to achieve more intelligent responses. Furthermore, compared to TBS methods, our think-while-speaking approach achieves comparable performance, with significantly lower CoT latency as analysed in Section \ref{sec:latency}.

To demonstrate the latency differences between the proposed method and baseline approaches, we select four models from Table \ref{tab:spokenmqa} that include a thinking process: Mini-Omni-Reasoner, MPS-tbs, MPS-thkfirst, and MPS-spkfirst. We calculate the number of extra tokens generated by the model from the end of the user's question to the generation of the first response token on Spoken-MQA. The results are shown in Table \ref{tab:latency}. It can be observed that compared to MPS-tbs, MPS-thkfirst achieves higher accuracy while exhibiting significantly lower response latency. Although the accuracy of MPS-spkfirst is slightly lower than that of MPS-tbs and MPS-thkfirst, its response is without latency. Furthermore, compared to Mini-Omni-Reasoner, which uses interleaved think and response tokens to achieve think-while-speaking, the proposed MPS methods achieve higher accuracy. Notably, the MPS-spkfirst attains this superior accuracy with zero latency. This indicates that MPS-spkfirst can play a more critical role in real-time dialogue scenarios with low-latency requirements. An example of the output of MPS-spkfirst is shown in Appendix \ref{sec:apx_exam}.

\begin{table}[t]
\caption{Performance of different methods on the URO-Bench. The evaluated approaches include: the direct response baseline without a thinking process (MPS-wo/\textit{thk}), Think-Before-Speaking (MPS-tbs), Think-First (MPS-thkfirst), and Speak-First (MPS-spkfirst). Results of baseline systems are taken from \cite{stepaudio2}. The results of \textit{Multilingual} of URO-Bench are included in \textit{English}. }
\label{tab:uro}
\renewcommand\arraystretch{1.1}
\begin{center}
\begin{tabular}{lccccccccc}
\toprule
\multirow{2}{*}{Method} & \multirow{2}{*}{Language} & 
  \multicolumn{4}{c}{Basic} & 
  \multicolumn{4}{c}{Pro} \\
  \cmidrule(lr){3-6} \cmidrule(lr){7-10} &
  & U. & R. & O. & Avg & U. & R. & O. & Avg
\\ \midrule
GPT-4o Audio & \multirow{5}{*}{Chinese} & 89.4& 65.5& 85.2& 78.6& 70.6& 57.2& 70.2& 67.1\\
GPT-Realtime & & 88.8 & 72.9 & 90.8 & 80.6 & 72.3 & 62.6 & 74.2 & 70.6 \\
Kimi-Audio & & 79.3& 64.7& 79.8& 73.6& 60.4& 59.3& 76.2& 66.0  \\
Qwen-Omni & & 59.7& 69.7& 77.3& 69.0& 59.0& 59.8& 58.7& 59.1 \\
Step-Audio 2 & & 91.1 & 75.5 & 86.1& 83.3& 74.8 & 63.2 & 65.1 & 68.3 \\
\midrule
MPS-wo/\textit{thk} & \multirow{4}{*}{Chinese}& 91.6 & 77.3& 87.7& 83.4 & 75.1 & 74.7 & 72.9 & 74.4 \\
MPS-tbs & & 92.6 & 82.4& 93.8 & 87.8 & 75.3 & 84.2 & 79.5 & 79.0\\
MPS-thkfirst & & \textbf{93.6} & \textbf{84.0} & \textbf{94.8} & \textbf{89.1} & 75.2 & 84.2 & \textbf{85.2} & \textbf{80.5}\\
MPS-spkfirst & & 92.5 & 82.5 & 93.1 & 87.6& \textbf{77.2} & \textbf{84.8} & 79.0& 79.9\\
\toprule
GPT-4o Audio & \multirow{5}{*}{English} & 90.2& 75.9& 90.4& 84.5& 60.7& 64.4& 78.5& 67.5\\
GPT-Realtime & & 87.4 & \textbf{84.1} & \textbf{94.1} & \textbf{88.1} & 59.7 & 74.5 & 76.1 & 68.9\\
Kimi-Audio & & 83.4& 42.3 & 60.4& 60.0& 50.3& 40.6& 56.0& 49.8 \\
Qwen-Omni & & 66.3& 69.6& 76.2& 70.6& 44.5& 63.9& 49.4 & 51.0 \\
Step-Audio 2 & & 92.7 & 76.5 & 84.9 & 83.9 & 64.9 & 67.8 & 66.3& 66.1 \\
\midrule
MPS-wo/\textit{thk} & \multirow{4}{*}{English} & 91.5 & 68.7 & 78.8 & 77.4& 73.4 & 79.2 & 55.8 & 65.1 \\
MPS-tbs & & 92.3& 81.5 & 87.5 & 86.1 & 76.4 & 86.4& 87.1 & 83.3 \\
MPS-thkfirst & & \textbf{94.2} & 81.4& 89.0 & 87.0& \textbf{76.5} & 89.3& \textbf{89.4} & \textbf{85.0}\\
MPS-spkfirst & & 94.1 & 78.5& 87.5 & 85.2 & 76.0 & \textbf{89.7}& 69.9 & 74.8\\
\bottomrule
\end{tabular}
\end{center}
\end{table}
\subsubsection{Evaluation on Speech-to-speech conversation}
Table \ref{tab:uro} shows the results of different methods on URO-Bench. It can be observed that MPS-thkfirst achieves higher performance than MPS-tbs on nearly all tasks and on average, under lower response latency. This may also be related to the implicit semantic alignment performed by MPS-thkfirst. Due to generating an initial response segment without prior thinking, MPS-spkfirst performs slightly worse than MPS-thkfirst, but still significantly outperforms the direct response method MPS-wo/\textit{thk}. Nevertheless, MPS-spkfirst features lowest response latency as analysed in Section \ref{sec:latency}, and its response performance remains close to or even better than that of MPS-tbs in some tasks, making it more suitable for scenarios requiring faster feedback. The experimental results demonstrate that the proposed MPS method maintains high performance on dialogue tasks, achieving performance comparable to TBS models while operating at significantly lower latency.

\section{Conclusion}
This paper proposes the MPS method, which enables SLMs to possess the ability to think while speaking. Inspired by the human thinking and response mechanism, we use a Formulation Brain LLM to continuously generate think segments, pacing the Articulation Brain LLM to utilize historical and current think segments, as well as historical responses, to generate current response segment, ensuring semantic coherence. Experimental results on mathematical reasoning and speech conversation tasks show that the proposed method significantly outperforms direct response methods and existing think-while-speaking methods. It achieves performance comparable or even better than methods that complete thinking before responding, while greatly reducing response latency. The proposed method breaks through the limitations of existing interleaved thinking and response-based think-while-speaking methods and provides an effective reference for researching real-time dialogue consistent with human thinking and response mechanisms. MPS also serves as the methodology behind our released Step-Audio R1.1 system.

\section{Acknowledgement}
We would like to express our sincere gratitude to Liang Zhao and Chengyuan Yao for their insightful suggestions and constructive discussions regarding the design of the model’s thinking mechanism. Their expertise greatly contributed to the development of the Formulation Brain LLM in this work.

\setlength{\bibsep}{0.5\baselineskip}
\bibliography{references}

\newpage
\appendix

\section{Appendix}
\subsection{Example of MPS-spkfirst}
\label{sec:apx_exam}

\begin{figure}[!htbp]
\begin{center}
\begin{minipage}[b]{0.89\linewidth}
  \centering
  \centerline{\includegraphics[width=18cm]{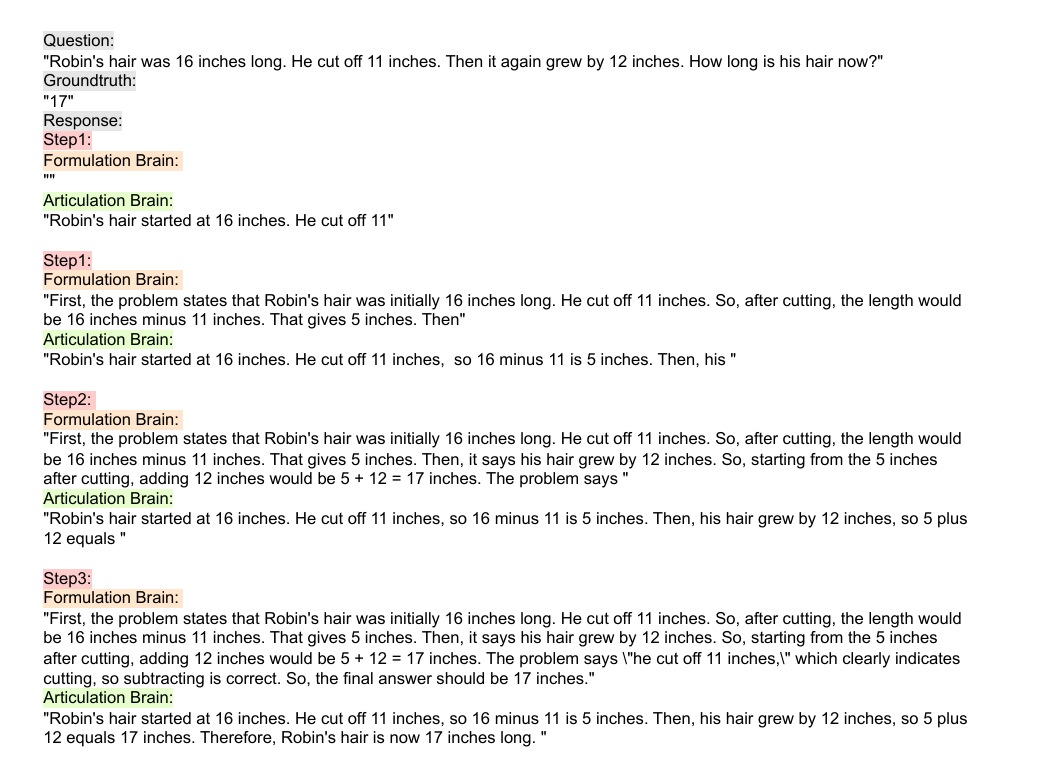}}
\end{minipage}
\vspace{-0.2cm}
\caption{An example of the output of MPS-spkfirst on the Spoken-MQA dataset. The Articulation Brain first generates a response segment. Simutaneously, Formulation Brain continuously generates new think segments, and each newly generated think segment is prefixed to the Articulation Brain, pacing it to generate new response segment.}
\label{fig:example}
\end{center}
\end{figure}

Figure \ref{fig:example} shows an example of MPS-spkfirst on Spoken-MQA. After receiving the user input, the Articulation Brain LLM first generates a response segment $\bm{S}^{res}_1$. Simultaneously, the Formulation Brain LLM produces the first think segment $\bm{S}^{thk}_1$. $\bm{S}^{thk}_1$ is then prefixed to the Articulation Brain LLM along with $\bm{S}^{res}_1$ to pace the Articulation Brain LLM in generating the second response segment $\bm{S}^{res}_2$. During this period, the Formulation Brain LLM generates the second think segment $\bm{S}^{thk}_2$. $\bm{S}^{thk}_2$ is further prefixed to the Articulation Brain LLM, where $\bm{S}^{thk}_1$, $\bm{S}^{thk}_2$, $\bm{S}^{res}_1$, and $\bm{S}^{res}_2$ collectively pace the Articulation Brain LLM to produce $\bm{S}^{res}_3$. This process repeats until the Formulation Brain LLM generates the complete think content, after which the Articulation Brain LLM continues generating content until completion.

\end{document}